\documentclass[letterpaper, 10 pt, conference]{ieeeconf}  % Comment this line out if you need a4paper
\usepackage{amsmath}
\usepackage{tensor}
\usepackage{bm}
\usepackage{amssymb}
\usepackage{graphicx}
\usepackage{caption}
\usepackage{color}
\usepackage{float}
\usepackage{rotating}
\usepackage{multirow}
\usepackage{hhline}
\usepackage{booktabs}
\usepackage{makecell}
\usepackage{longtable}
\usepackage{array}

\usepackage{paralist}
\usepackage{multicol}
\usepackage{chngcntr}
\usepackage{arydshln,leftidx,mathtools}
\usepackage{threeparttable}
\usepackage{tabularray}
\usepackage[noadjust]{cite}

\IEEEoverridecommandlockouts                              % This command is only needed if 
\title{\LARGE \bf
Source-Free Bistable Fluidic Gripper for Size-Selective and Stiffness-Adaptive Grasping}

\author
{Zhihang Qin†, Yueheng Zhang†, Wan Su, Linxin Hou, Shenghao Zhou, Zhijun Chen, \\Yu Jun Tan* and Cecilia Laschi*
\thanks{This work was supported by the bridging fund (AI-Driven Soft Robots for Marine and Unstructured Environments), the start-up grant RoboLife (Soft Robots with morphological adaptation and life-like abilities), DESTRO (Dextrous, strong yet soft robots), ITALY–SINGAPORE grant, MAE (Italy) and A*STAR (Singapore), Grant \#R22I0IR124, and REBOT (Rethinking underwater robot manipulation), Ministry of Education, Singapore, Grant \#T2EP50221-0028. (Corresponding author: Yu Jun Tan, Cecilia Laschi)}
\thanks{Zhihang Qin, Yueheng Zhang, Wan Su, Longxin Kan, Shenghao Zhou, Yu Jun Tan and Cecilia Laschi are with Department of Mechanical Engineering, National University of Singapore, Singapore, Singapore}
\thanks{Zhihang Qin, Shenghao Zhou, Yu Jun Tan and Cecilia Laschi are with the Advanced Robotics Centre, National University of Singapore, Singapore, Singapore}
\thanks{Linxin Hou is with Department of Electrical and Computer Engineering, National University of Singapore, Singapore, Singapore}
\thanks{† Zhihang Qin and Yueheng Zhang contributed equally to this work.}
}

\begin{document}
\maketitle
\thispagestyle{empty}
\pagestyle{empty}

% Abstract-------------------------------------------------------------
\begin{abstract}

Conventional fluid-driven soft grippers typically depend on external sources, which limit portability and long-term autonomy. This work introduces a self-contained soft gripper with fixed size that operates solely through internal liquid redistribution among three interconnected bistable snap-through chambers. When the top sensing chamber deforms upon contact, the displaced liquid triggers snap-through expansion of the grasping chambers, enabling stable and size-selective grasping without continuous energy input. The internal hydraulic feedback further allows passive adaptation of gripping pressure to object stiffness. This source-free and compact design opens new possibilities for lightweight, stiffness-adaptive fluid-driven manipulation in soft robotics, providing a feasible approach for targeted size-specific sampling and operation in underwater and field environments.

\end{abstract}

% Introduction_New---------------------------------------------------

\section{INTRODUCTION}\label{section:1}
Fluid-driven soft mechanical systems, which encompass pneumatic and hydraulic architectures, have attracted significant attention for their compliance, adaptability, and energy-dissipating capabilities. By modulating internal pressure and fluid flow, they are applied in various fields for controlled deformation, motion, and damping. In biomedical engineering, they enable flexible microfluidic devices to perform precise liquid manipulation and rehabilitation exoskeletons to achieve lightweight, conformal, and safe human–machine interfaces \cite{nie_2020_hydrogels, zhu_2022_soft, cianchetti_2018_biomedical}. In industrial and civil engineering, fluid-filled damping layers and viscous channels dissipate mechanical energy for impact protection and seismic resistance \cite{sheikhi_2023_vibration}.

\begin{figure}[!t]
    \centering
    \includegraphics[width=1\columnwidth]{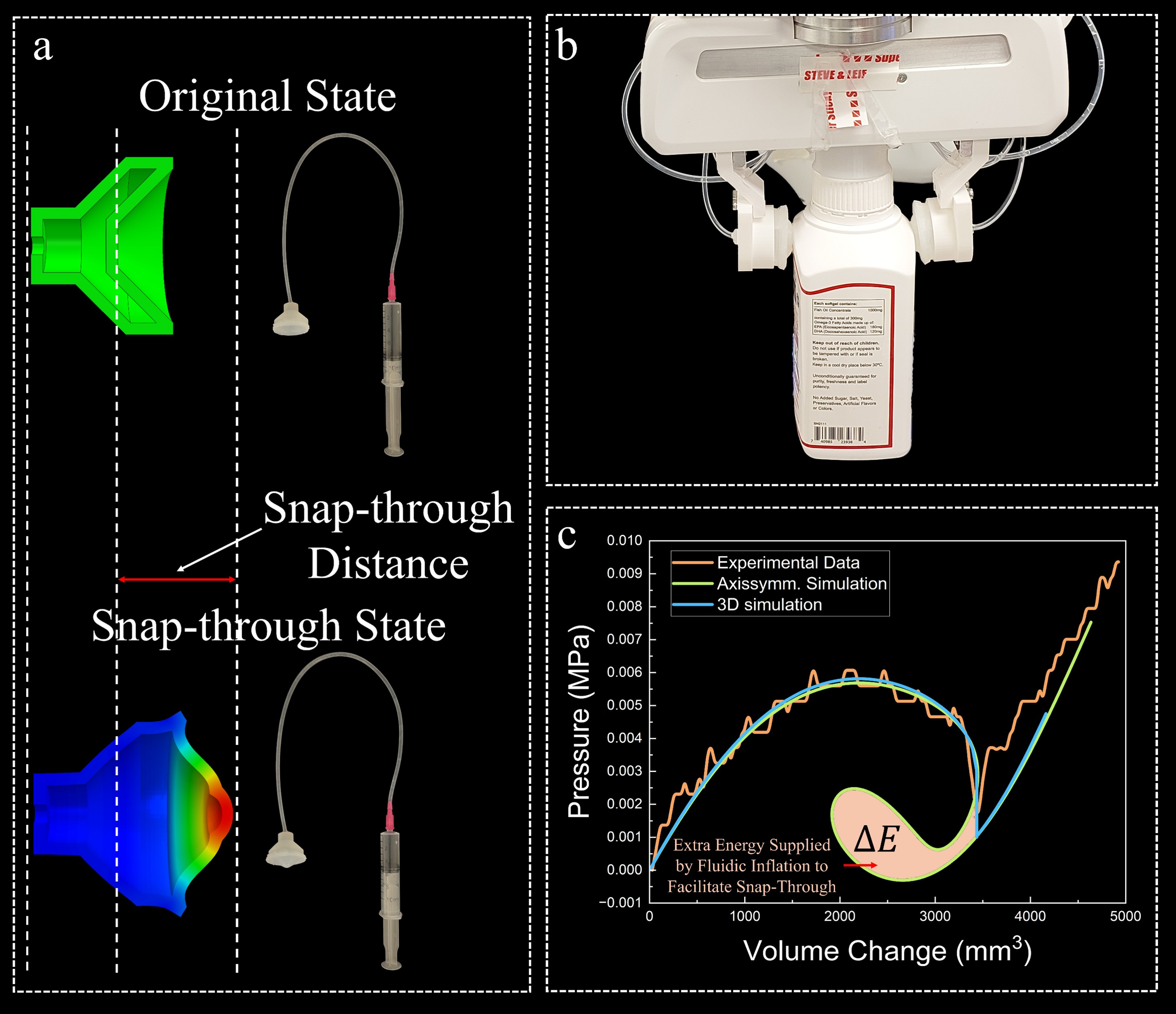} 
    \caption{ The pictures showing the working mechanism of the self-sustaining fluid-driven gripper. (a) The original state and snap-through state of the fundamental unit of the self-sustaining fluid-driven gripper. (b) The complete gripper structure that utilizes a snap-through response to hold the fish oil bottle.
    (c) The data curves that illustrate the simulation curve of pressure-volume change along equilibrium path (Axissymm. Simulation), simulation curve of pressure-volume change under snap-through response (3D Simulation), and experimental curve of pressure-volume change under snap-through response (Experimental Data).
    }
    \label{fig:graphical_abstract}
\end{figure}

In addition to these domains, fluidic actuation has become one of the dominant strategies for compliant and adaptive manipulation \cite{laschi_2016_soft, laschi_2025_soft, xin_2023_the}. Fluid-driven soft grippers employ elastomeric chambers that bend \cite{polygerinos_2017_soft}, expand \cite{hao_2021_a}, or contract \cite{koivikko_2021_3dprinted} under pressure to conformally grasp objects of various shapes and stiffness \cite{kanlongxin_2024_bistable, zhihang_2025_edible}. Their continuum deformation enables safe interaction with fragile targets \cite{li_2023_bioinspired, zhang_2025_soft}. However, most existing systems rely on bulky external pumps, compressors, or high-pressure cylinders to maintain actuation. Even portable gas cylinders offer limited endurance as compressed gas depletes over repeated cycles, resulting in tethered, heavy, or short-lived setups \cite{jung_2024_untethered, li_2022_soft, mcdonald_2021_hardware, youssef_2022_underwater}. These limitations hinder portability and autonomy, particularly in underwater or field-deployed robots, where system compactness and energy efficiency are critical \cite{kelasidi_2015_energy, fang_2022_a, shen_2020_an, rus_2015_design}.

\begin{figure}[!t]
    \centering
    \includegraphics[width=1\columnwidth]{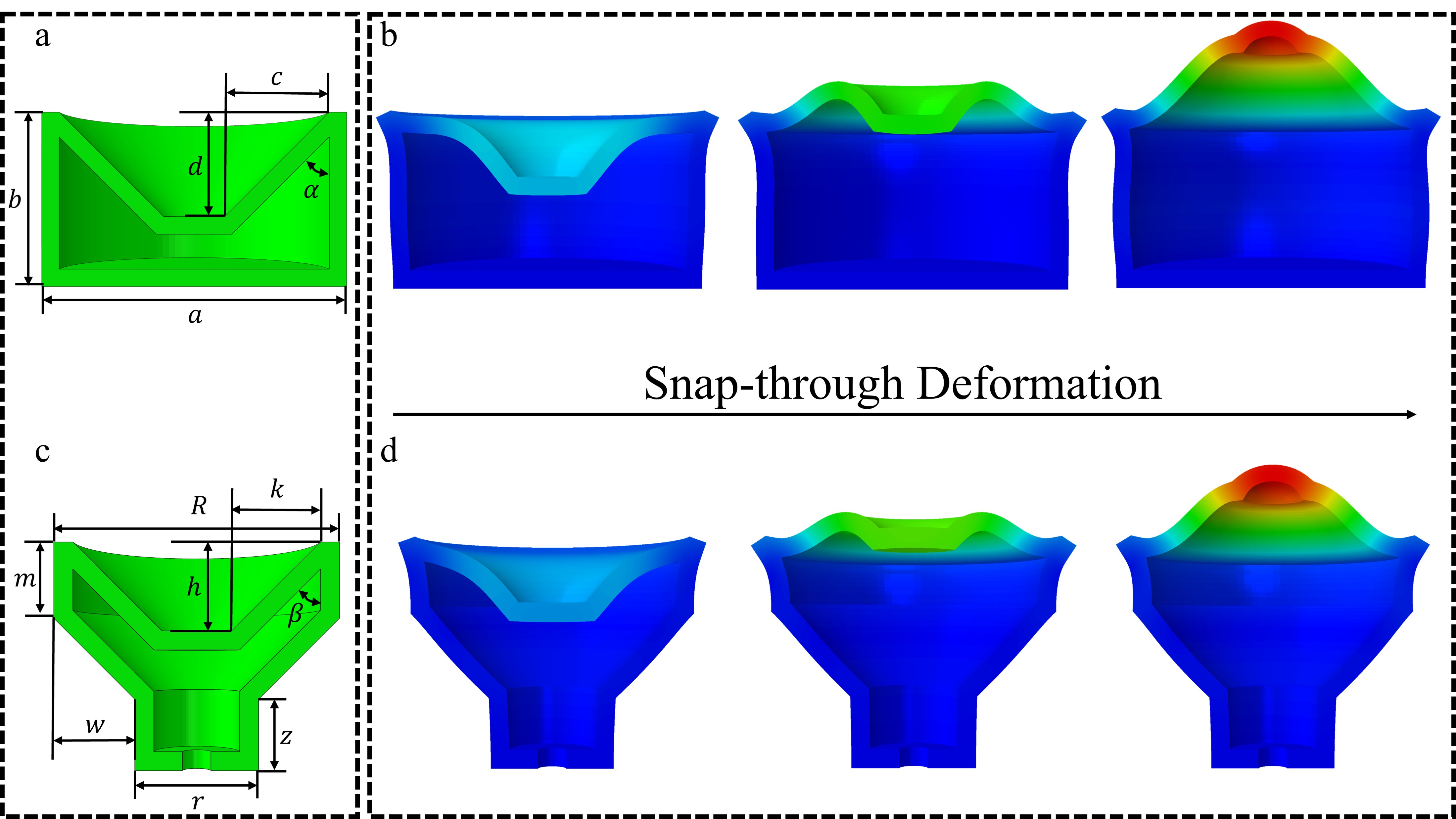}
    \caption{Modeling and simulations. (a, c) 3D modeling of the contact chamber and gripping chamber. (b, d) Numerically obtained 3D deformation processes of the contact chamber and gripping chamber in general-static analysis.}
    \label{fig:design}
\end{figure}

Although such limitations remain, they have inspired the exploration of potential structural strategies to address them. Bistable snap-through structures, as one representative example, have emerged as a key paradigm for achieving shape retention without continuous energy input while maintaining rapid state transitions \cite{thuruthel_2020_a, li_2025_design, chi_2022_bistable}. Their double local minimal energy points allow the structure to remain in a deformed configuration once triggered, and the snap-through instability enables rapid, energy-efficient switching between stable states \cite{vasios_2021_universally, librandi_2021_programming, tang_2024_bistable, gorissen_2020_inflatable, zareei_2020_harnessing}. In contrast, conventional fluid-driven actuators relax immediately when pressure is released, losing their grip. Integrating bistability into fluid-driven systems thus provides a promising route toward stable, energy-saving, and responsive operation.

Building on the bistable mechanism, we propose a self-contained, source-free fluidic gripper with a fixed grasping size, which operates through internal liquid redistribution among interconnected bistable snap-through chambers. This design eliminates the need for bulky or short-lived external pumps and enables stable operation within a closed liquid circuit. As shown in Fig. \ref{fig:graphical_abstract}, when the top contact chamber touches an object, the displaced liquid drives snap-through deformation in the lateral gripping chambers, achieving grasping without continuous energy input.

The fixed-distance configuration allows reliable interaction between the contact chamber and the object, triggering a snap-through grasping response when the object size falls within the appropriate range, while objects larger or smaller than this range are effectively filtered out. Experimentally, the gripper successfully grasped objects when the preset distance between the clamping fixtures was fixed at 4 mm (narrow range) and 10 mm (wider range). Furthermore, objects of different stiffness impede the motion of the sensing chamber to varying degrees, generating distinct internal hydraulic responses and resulting in passive stiffness-adaptive gripping forces.

Overall, this mechanism enables a source-free, size-selective, and stiffness-adaptive gripper, offering a potential approach for self-adaptive manipulation and size-specific sampling in underwater and field environments.

% Section3-----------------------------------------------------------
\section{Design and Simulation}\label{section:2}

\subsection{Geometry of Bistable Chamber} 

The fundamental structure of our gripper design consists of two gripping chambers used for grasping and a contact chamber used for sensing objects. The geometry and design parameters of these two different types of chambers are shown in Fig. \ref{fig:design}a and c, featuring the same bistable configuration.

\begin{figure*}[!t]
    \centering
    \includegraphics[width=1\textwidth]{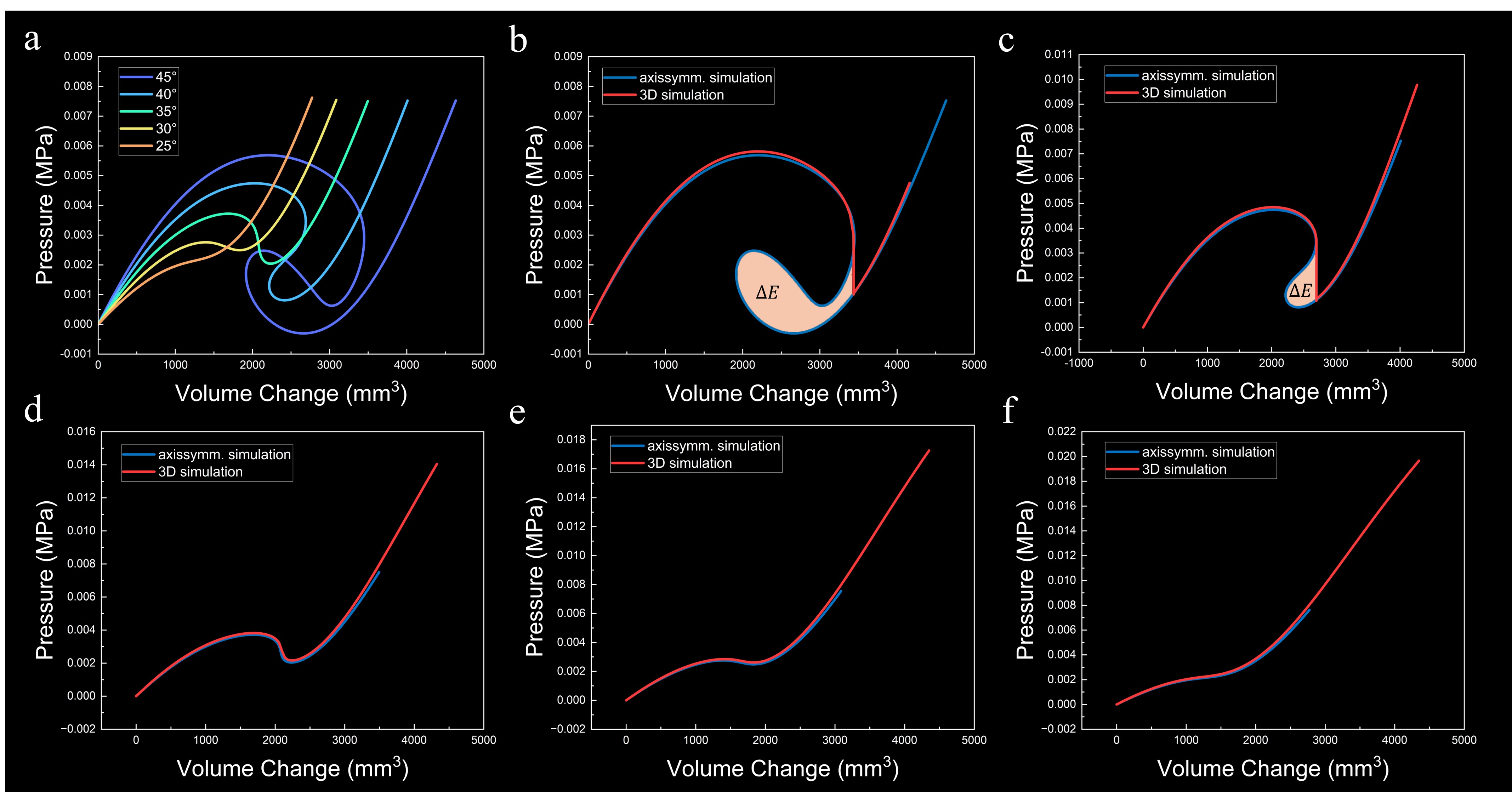}
    \caption{Pressure–volume response of the snap-through membranes with different tilt angles in gripping chambers along the equilibrium and snap-through paths. (a) Pressure–volume curves of snap-through membranes with various tilt angles along the equilibrium path. (b–f) Comparison of the pressure–volume curves between the equilibrium and snap-through paths for tilt angles of $45^{\circ}$, $40^{\circ}$, $35^{\circ}$, $30^{\circ}$, and $25^{\circ}$, respectively.}
    \label{fig:simulation-for-design}
\end{figure*}

In Fig. \ref{fig:design}a, the wall thickness of the contact chamber is fixed at 2 mm, and the design parameters $a$ = 35 mm, $b$ = 20 mm, and $c$ = 12 mm remain constant. Only the tilt angle $\alpha$ and its associated design parameter $d$ vary with changes in the angle value.

In Fig. \ref{fig:design}c, the upper and lower inclined surfaces of the gripping chamber are kept parallel. The wall thickness of the gripping chamber is also fixed at 2 mm, and the design parameters $R$ = 30 mm, $r$ = 13 mm, $w$ = 8.5 mm, $m$ = 8 mm, and $z$ = 7.5 mm remain constant. Only the tilt angle $\beta$ and its associated design parameters $h$ and $k$ vary with changes in angle value.

Fig. \ref{fig:design}b and d illustrate the snap-through deformation process of the contact chamber and gripping chamber, respectively.

The reason for fixing multiple design parameters rather than varying all of them is that the bistable characteristics of the snap-through membrane are strengthened with increasing tilt angle but weakened as other parameters controlling the membrane diameter increase. Introducing too many control variables would complicate the design process and hinder clear analysis. Therefore, in this work, we focus on the primary parameter influencing the bistable behavior of the snap-through membrane, the tilt angle, to determine the final design configuration.

\subsection{Finite Element Analysis Setup} 

Finite element simulations were conducted in ABAQUS 2022/Standard to analyze the mechanical response of the snap-through chamber and optimize its tilt angle. The chamber exhibits a highly nonlinear mechanical response along its equilibrium path due to structural instability. Consequently, its load–deformation curve typically contains one or more limit points, where the tangent stiffness approaches zero. This non-monotonic behavior prevents Newton’s method, commonly employed in the general static or dynamic implicit solver of ABAQUS, from accurately tracing the equilibrium path. Instead, the solver can only capture the response along the snap-through branch, failing to reproduce the true triggering mechanism. To address this limitation, a 2D axis-symmetric analysis using the ABAQUS Riks solver was additionally performed to accurately compute the mechanical response throughout the post-buckling regime.

In all analyses, the response of the silicone rubber used to fabricate the snap-through chambers (Smooth-On, Inc. | Ecoflex 00-30, Dragon skin 00-30, and Sil950) is modeled using a Neo-hookean material model with strain energy density function $W$ given by

\begin{equation}
    W = C_{10} (I_{1} J^{-\frac{2}{3}} - 3) + D_{1}(J - 1)^{2},
    \label{eq:neo}
\end{equation}

where $C_{10}$ and $D_{1}$ are the hyperelastic material parameters, $I_{1} = \mathrm{tr}(F^{\top} F)$ is the first principal invariant, and $J$ is the determinant of the deformation gradient tensor $F$, which is the ratio of the deformed volume to the original volume. According to the manufacturer datasheets provided by Smooth-On and assuming the silicone rubbers to be incompressible, the mechanical responses of Ecoflex 00-30, Dragon Skin 00-30, and Sil950 were modeled using the parameters $(C_{10}, D_{1}) = (0.0115\ \mathrm{MPa}, 0)$, $(0.1\ \mathrm{MPa}, 0)$, and $(0.4\ \mathrm{MPa}, 0)$, respectively.

\begin{figure}[!t]
    \centering
    \includegraphics[width=1\columnwidth]{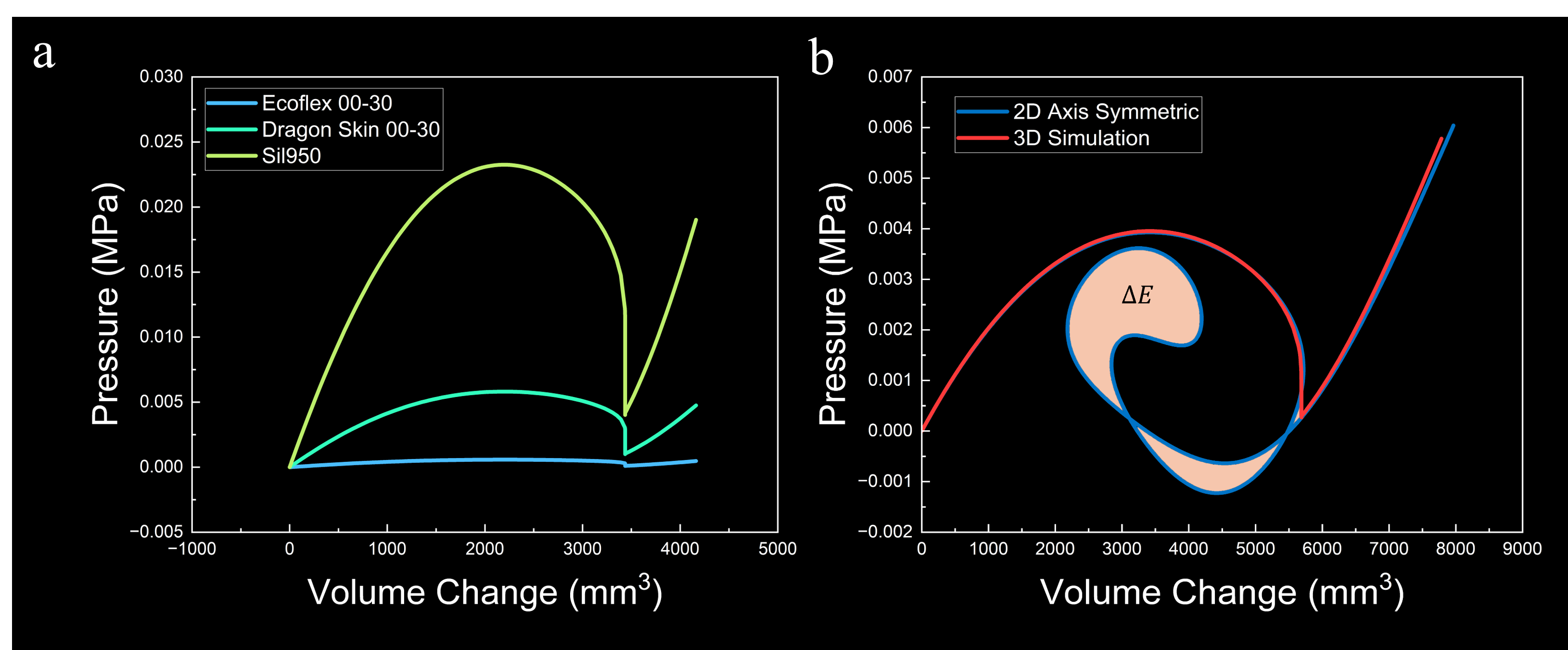} 
    \caption{Pressure–volume response of the snap-through membranes with a $45^{\circ}$ tilt angle in the gripping and contact chambers. (a) Effect of material variation on the pressure–volume response. (b) Pressure–volume response of the contact chamber along the equilibrium and snap-through paths.
    }
    \label{fig:simulation-for-design-2}
\end{figure}

In the following sections, we present the different types of simulation setups implemented to analyze the mechanical behavior of snap-through chambers under quasi-static hydraulic inflation.

\subsubsection{3D Simulation Setup}

We first conducted a full three-dimensional finite element simulation of the snap-through chambers. A 3D model was constructed and discretized using a non-structured mesh of 8-node linear hexahedral elements (ABAQUS element type: C3D8H). The mesh density was adjusted to ensure that at least four elements discretized the thickness of the thinnest wall. To eliminate rigid body translation and rotation, a zero vertical displacement boundary condition ($u_y = 0$) was applied at the bottom surface. All models were inflated through interaction with a hydraulic fluid (density $\rho$ = 1000 $\mathrm{kg/m^3}$, bulk modulus $K$ = 2000 $\mathrm{MPa}$) contained within the internal cavity. The volume-controlled hydraulic inflation was implemented by simulating the virtual thermal expansion of the hydraulic fluid. “Virtual” indicates that this is not an actual thermal expansion. The snap-through chamber is not assigned any material parameters coupled with a thermal field, ensuring that this operation serves solely to realize volume-controlled hydraulic inflation. This hydraulic inflation process was related to the change in cavity volume $\Delta V$ through

\begin{equation}
    \frac{\Delta V}{V_{0}} = 3 \alpha_{T} \Delta T,
    \label{eq:volume-expansion}
\end{equation}

where $\Delta T$ is the change in temperature, $\alpha_T$ is the coefficient of thermal expansion of the fluid, and $V_0$ is the initial cavity volume. In the simulations, we set $\alpha_T = 0.0034\ \mathrm{m/(m\cdot K)}$ and gradually increased the temperature $T$ until isochoric snap-through occurred. The hydraulic inflation process was simulated using the dynamic implicit solver, with a silicone rubber density of $\rho = 1000\ \mathrm{kg/m^3}$. Quasi-static conditions were ensured by using a total simulation time of 1 s, a minimum increment size of $1\times10^{-10}$ s, a maximum increment size of 0.01 s, and a maximum of 10000 increments, which minimized the kinetic energy of the model.

\subsubsection{axis-symmetric Simulation Setup}

While the 3D finite element (FE) simulations accurately capture and predict the snap-through response of the chambers, their high computational cost limits their efficiency in post-buckling analysis along the equilibrium path. To address this, the actuator deformation was assumed to be axis-symmetric, and the models were discretized using 4-node bilinear axis-symmetric solid elements (ABAQUS element type: CAX4H). The mesh size was adjusted to ensure that at least four elements discretized the thickness of the thinnest wall. To evaluate the energy released during isochoric snap-through, the full pressure–volume relationship was obtained using the Riks solver implemented in ABAQUS. Similar to the 3D simulations, the axis-symmetric models were inflated via a fluid–structure interaction with a hydraulic fluid (density $\rho = 1000\ \mathrm{kg/m^3}$, bulk modulus $K = 2000\ \mathrm{MPa}$) and hydraulic inflation was terminated when the pressure $p$ reached

\begin{equation}
    p = 1.5p_s,
    \label{eq:critical}
\end{equation}

where $p_s$ is the critical pressure corresponding to the onset of snap-through. The factor of 1.5 accounts for the fact that most chambers are not thin-walled (i.e., $R/t < 25$). With respect to the boundary condition setup, apart from the axis-symmetric constraint, all other boundary conditions and the volume-controlled hydraulic inflation settings are consistent with those used in the 3D simulation setup.

\subsection{Simulation Analysis of Snap-through Chambers} 

The core of the bistable snap-through chamber design lies in two key criteria. First, the two response curves corresponding to the equilibrium and snap-through paths should exhibit an enclosed area within a specific range. Second, the pressure–deformation response should contain negative pressure values. The area difference between the two curves represents the additional mechanical energy input from the applied load, which facilitates rapid snap-through deformation. The presence of negative pressure values indicates a reduction in strain energy within that region, confirming that the chamber exhibits bistable characteristics under the given design parameters.

As shown in Fig. \ref{fig:simulation-for-design}a, the tilt angle has a significant influence on the mechanical response of the gripping chamber along the equilibrium path. As the tilt angle decreases, the tendency for snap-through behavior progressively weakens. At $20^{\circ}$, the number of limit points in the load–deformation curve of the gripping chamber is markedly reduced compared with $45^{\circ}$, and the valley region of the curve becomes noticeably flatter. Furthermore, as illustrated in Fig. \ref{fig:simulation-for-design}b–f, a closed area enclosed by the two response curves corresponding to the equilibrium and snap-through paths appears only when the tilt angle is greater than or equal to $40^{\circ}$. Negative pressure values are observed exclusively at $45^{\circ}$. These results indicate that only tilt angles of $45^{\circ}$ or higher satisfy the bistable design requirements of the snap-through chamber. Considering that excessive snap-through displacement should be avoided during object grasping to prevent over-deformation and potential damage to the target, a tilt angle of $45^{\circ}$ was selected as the optimal design parameter.

\begin{figure}[!t]
    \centering
    \includegraphics[width=1\columnwidth]{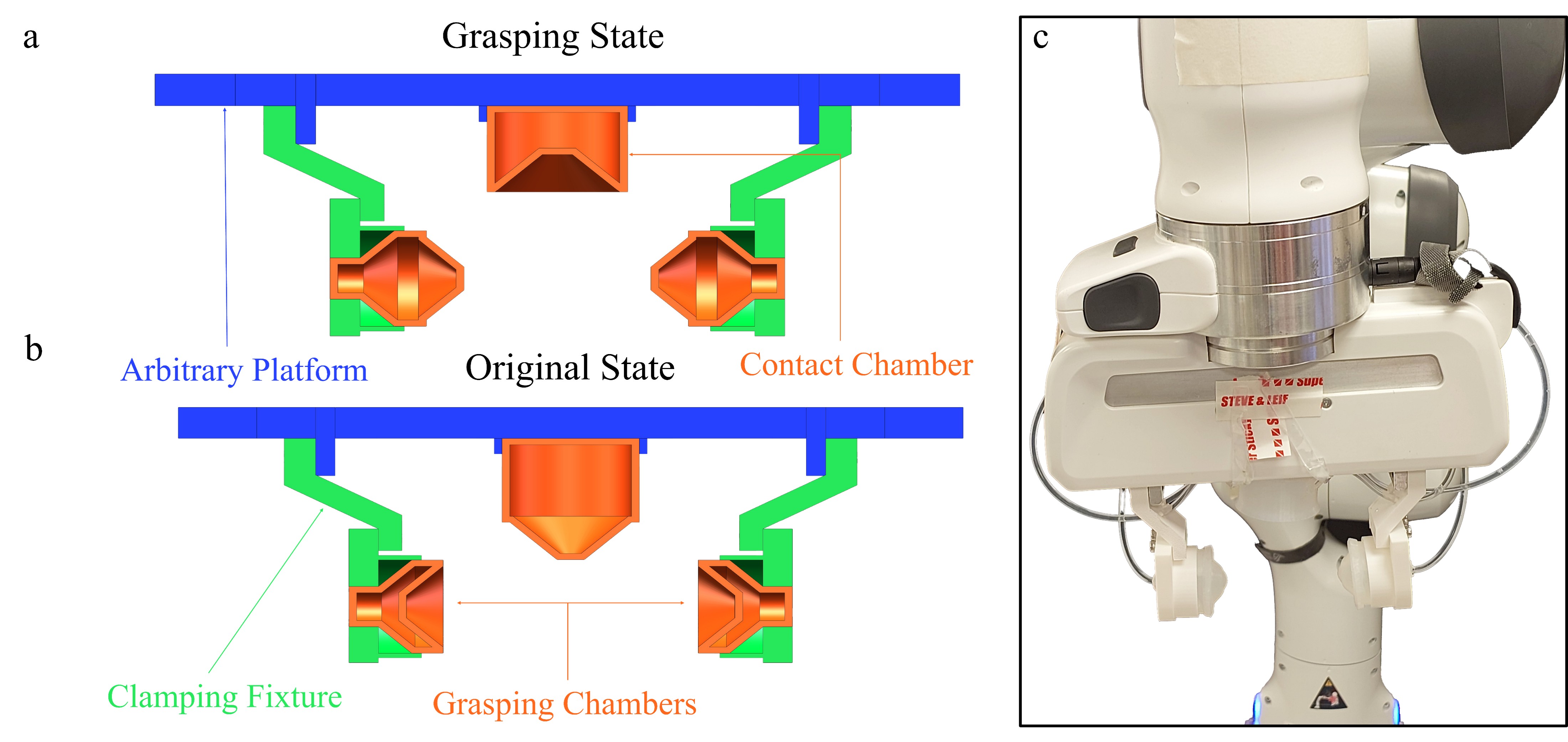} 
    \caption{Design and experimental setup of the gripping module. (a–b) Schematic diagrams showing the overall configuration of the fixed-size gripping module. The three orange chambers correspond to the contact chamber and two gripping chambers; the green components represent clamping fixtures used to attach the module to supporting structures, and the blue component denotes a generic platform (e.g., robotic arm or mobile base) capable of carrying the module. (a) Grasping state; (b) initial undeformed state. (c) Photograph showing the gripping module mounted on the robotic arm for proof-of-concept experiments.
    }
    \label{fig:module-design}
\end{figure}

After determining the optimal tilt angle of the gripping chamber, the structural design was finalized. A full three-dimensional analysis was then implemented to evaluate the mechanical response of the gripping chamber using silicone rubbers of different stiffness. As shown in Fig. \ref{fig:simulation-for-design-2}a, the elastic modulus of the material does not alter the qualitative characteristics of the mechanical response but increases the pressure threshold required to trigger snap-through.

Since the contact chamber shares the same snap-through configuration as the gripping chamber, its tilt angle was also set to $45^{\circ}$. As illustrated in Fig. \ref{fig:simulation-for-design-2}b, the mechanical response curves of the contact chamber, modeled with the material parameters of Dragon Skin 00-30, fully satisfy the two key criteria outlined at the beginning of this subsection, the presence of an enclosed area between the equilibrium and snap-through paths and the occurrence of negative pressure values. Therefore, the designs of both chambers were finalized and adopted for subsequent experimental validation and demonstration.

\begin{figure}[!t]
    \centering
    \includegraphics[width=1\columnwidth]{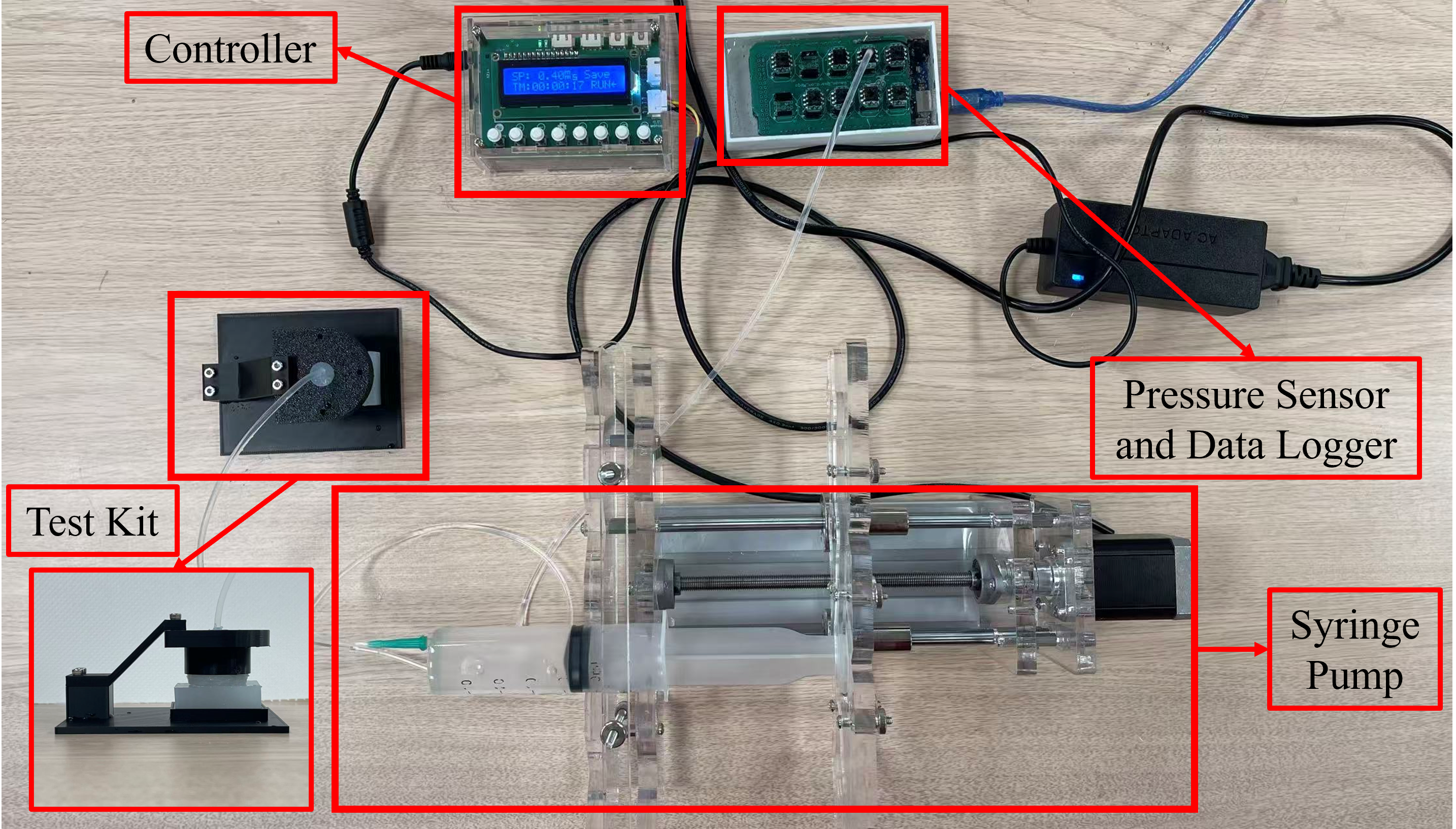} 
    \caption{The picture shows the experimental setup used to characterize the mechanical response of the chamber. It consists of an injection pump, a controller, pressure sensors, a data logger, and a custom fixture for securing the snap-through chamber.
    }
    \label{fig:test-kit}
\end{figure}

\begin{figure}[!t]
    \centering
    \includegraphics[width=1\columnwidth]{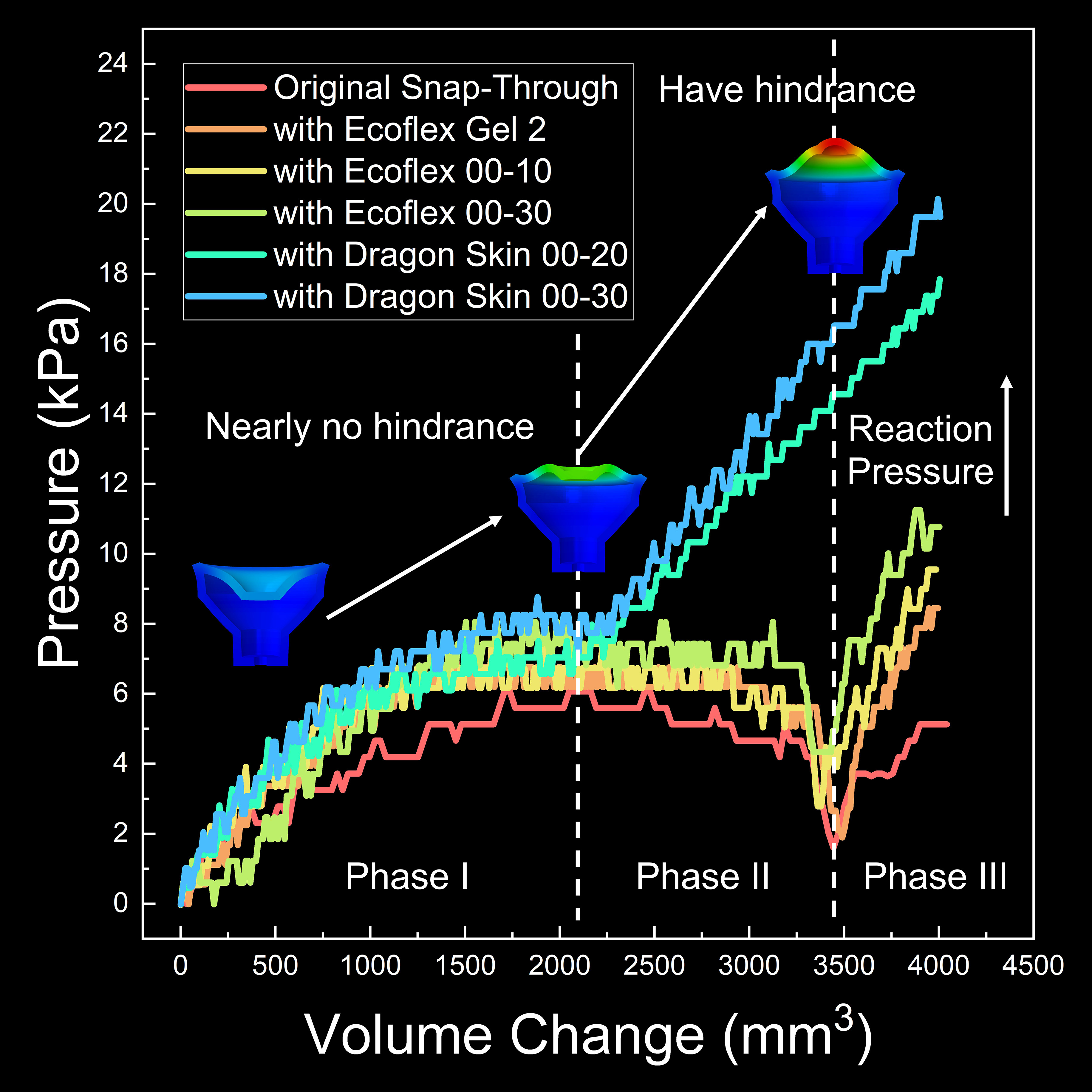} 
    \caption{Quasi-static mechanical response of the gripping chamber fixed in a custom fixture, tested without obstruction and with silicone blocks of different stiffness. When an object is present, the chamber and block are initially separated by a small gap, remaining on the verge of contact. The snap-through process consists of three stages: (I) pre-contact deformation, (II) contact and completion of snap-through, and (III) post-snap-through pressurization by continued water injection.
    }
    \label{fig:test-result-1}
\end{figure}

% Section3----------------------------------------------------------
\section{Experiment and Analysis}

\subsection{Module Fabrication}\label{subsection:mf}

As shown in Fig. \ref{fig:module-design}, the experimental gripping module consists of two main parts: two rigid clamping fixtures, which secure the flexible structure and connect it to the platform, and three flexible chambers—two gripping chambers and one contact chamber—responsible for object sensing and grasping. The rigid clamping fixtures were fabricated via FDM 3D printing using a Bambu Lab X1C printer, while the three silicone flexible chambers were produced through mold casting. To balance the snap-through triggering threshold, we selected Smooth-On Dragon Skin 00-30 as the casting material. A lower threshold would make the gripping module overly sensitive, resulting in insufficient gripping force and potential object slippage, whereas a higher threshold would make snap-through activation difficult and could damage the object due to excessive gripping force. 

For the experiments of the mechanical response of a single snap-through chamber under quasi-static conditions, a silicone tube (outer diameter: 3 mm, inner diameter: 2 mm) was used to directly connect the chamber to the syringe pump (Fig. \ref{fig:test-kit}). 

For the gripping experiments, the contact chamber and the two gripping chambers were first filled with water, ensuring that the contact chamber was in its snap-through state while the gripping chambers remained in their initial undeformed state. The chambers were then mounted onto the clamping fixture, interconnected using silicone tubes of identical dimensions, and sealed with a silicone-based epoxy adhesive to prevent leakage. The assembled module was subsequently integrated with a robotic arm (Franka Emika robot) via the clamping fixture (Fig. \ref{fig:design}) and employed as a soft gripper for proof-of-concept grasping demonstrations.

\subsection{Mechanical Response Testing Using Objects with Different Stiffness}

The mechanical responses of the contact and gripping chambers along the equilibrium and snap-through paths without object contact were analyzed in Section \ref{section:2}. However, during actual gripping, contact with an object impedes the snap-through motion, while the clamping fixture that restrains the gripping chamber may also undergo varying degrees of deflection depending on the stiffness of the object. These factors lead to mechanical responses that differ substantially from the free snap-through behavior observed under hydraulic inflation. To investigate how the gripping chamber’s mechanical response is affected by obstruction from objects of different stiffness, the experimental setup shown in Fig. \ref{fig:test-kit} was constructed.

To standardize contact conditions with the test objects, silicone rubbers of different stiffness (Ecoflex Gel 2, Ecoflex 00-10, Ecoflex 00-30, Dragon Skin 00-20, and Dragon Skin 00-30) were uniformly cast into blocks of identical size. The contact gap between the gripping chamber and each block was fixed at 2 mm using a custom fixture to ensure consistent experimental conditions. The syringe pump (purchased from Alibaba) was operated at a piston advancement speed of 0.2 mm/s, corresponding to a volumetric flow rate of 129.385 $\mathrm{mm^3/s}$ for a syringe with an inner diameter of 28.7 mm. Throughout the experiments, data acquisition and transmission were carried out using a sensing module comprising a liquid-compatible pressure sensor (First Sensor SQ273-P001GZ8P) and an Arduino Mega 2560 module.

Fig. \ref{fig:test-result-1} shows the quasi-static mechanical response of the gripping chamber, which was fixed in a custom fixture for controlled testing. The chamber was tested both in the absence of any obstruction and with silicone blocks of different stiffness placed in front of it. These blocks were made from silicone casting materials of varying stiffness, representing objects of different stiffness. When an object was present, the gripping chamber and the block were positioned with a small initial gap of approximately 2 mm, ensuring they remained on the verge of contact before hydraulic inflation.

During the volume-controlled hydraulic inflation process, the snap-through path of the chamber could be divided into three distinct phases. In the first phase, the snap-through membrane deformed without direct contact with the object. In the second phase, the membrane came into contact with the block and completed the snap-through transition. In the third phase, after the snap-through event was completed, additional water was injected into the chamber by the syringe pump, further increasing the internal volume.

\begin{figure}[!t]
    \centering
    \includegraphics[width=1\columnwidth]{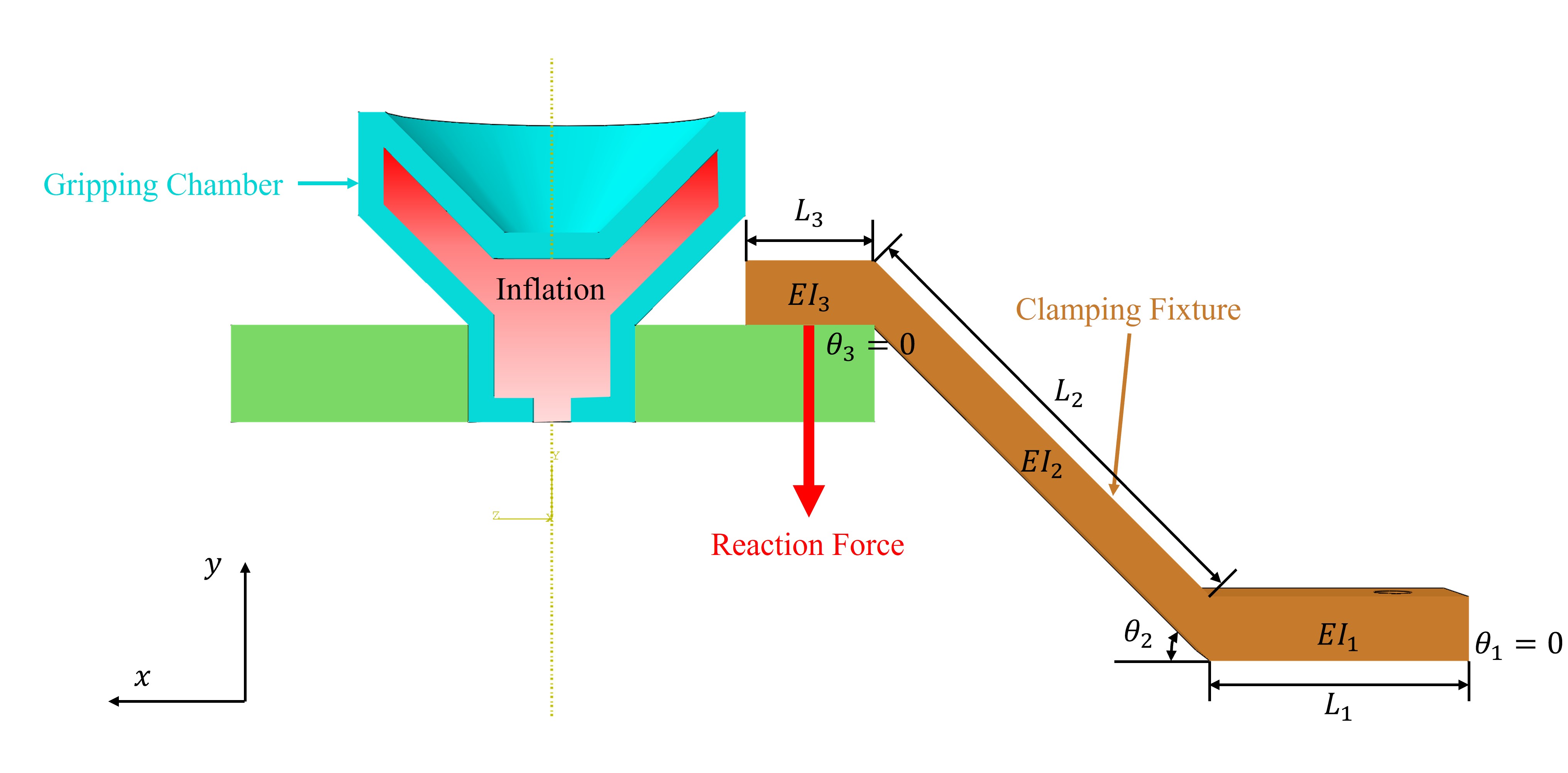} 
    \caption{Schematic diagram of the single-sided gripper module setup. The blue component represents the fluidic gripping chamber. The green and orange parts together form the clamping fixture. The inflation-induced force is transmitted through screws connecting the green and orange components, applying a concentrated load P on the orange component.
    }
    \label{fig:gripper_structure}
\end{figure}

\begin{figure}[!t]
    \centering
    \includegraphics[width=1\columnwidth]{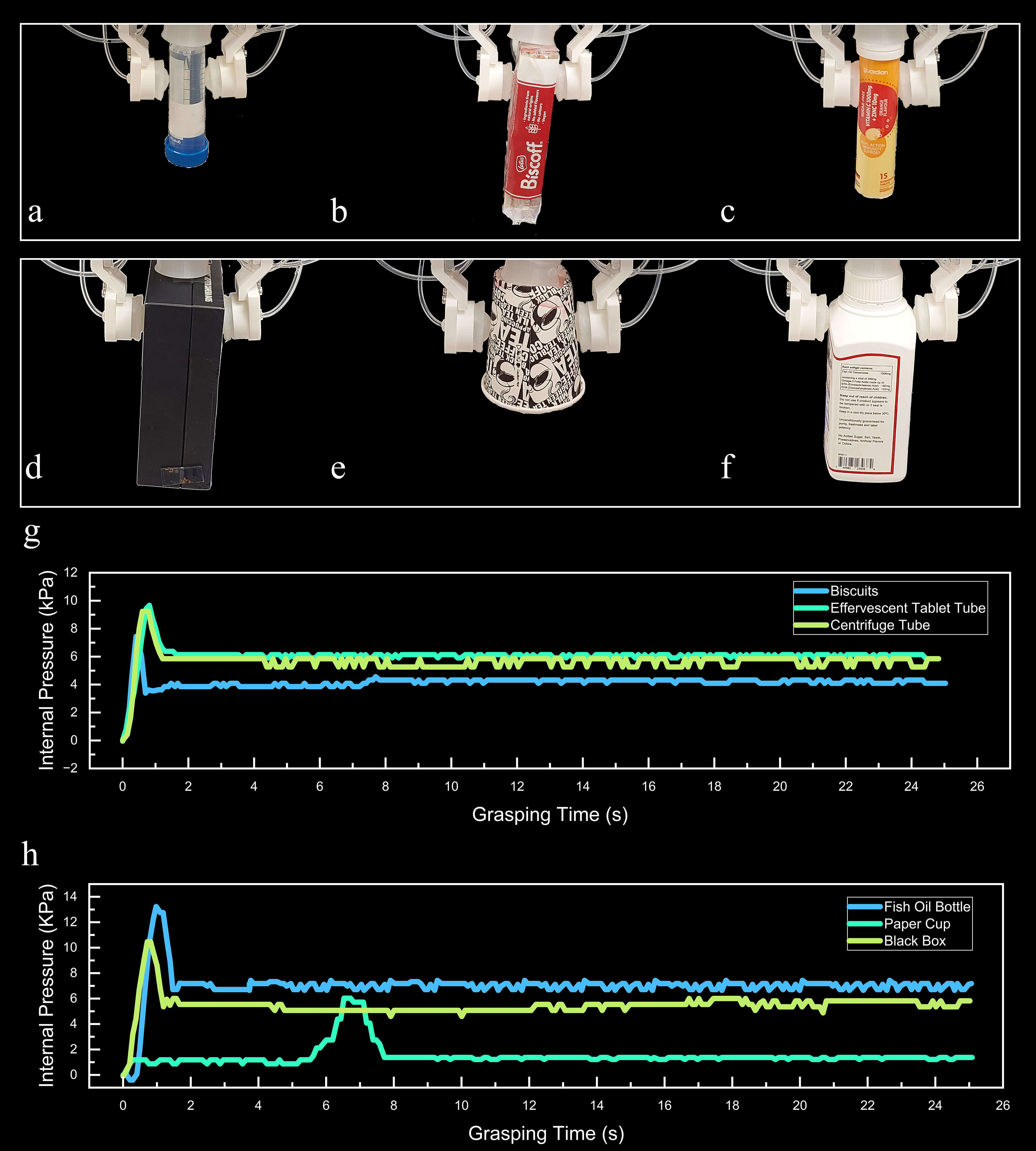} 
    \caption{Objects grasped and corresponding reaction hydraulic pressures during the grasping process: (a) centrifuge tube, (b) biscuits with flexible packaging, (c) effervescent tablet tube, (d) empty black box, (e) water-soaked paper cup, (f) fish oil bottle, (g) reaction hydraulic pressure of the small-size group, and (h) reaction hydraulic pressure of the large-size group.
    }
    \label{fig:test-result-2}
\end{figure}

These three phases clearly demonstrate that the presence and stiffness of the object have a pronounced influence on the latter two phases of the snap-through process, significantly affecting the chamber’s pressure evolution and deformation behavior. Under fixed-distance response and volume-controlled hydraulic inflation conditions, the gripping chamber passively regulates its internal hydraulic pressure according to the stiffness of the grasped object, aided by the deflection change of the clamping fixture. As a result, the module generates different output pressures—and thus different gripping forces—when grasping soft and hard objects. The reaction pressure reaches approximately 14 kPa for stiff blocks and 2–4 kPa for soft blocks, at a hydraulic inflation volume of about 3500 \text{$\mathrm{mm^3}$}, corresponding to the point at which the gripping chambers complete snap-through and just enter phase III.

\subsection{Theoretical Analysis of the Stiffness-Adaptive Mechanism aided by the Clamping Fixture}

In addition to the experimental results, a theoretical analysis based on the Euler--Bernoulli beam formulation for the clamping fixture (a three-segment piecewise linear beam) is provided to elucidate the underlying stiffness-adaptive mechanism.

As shown in Fig.~\ref{fig:gripper_structure}, the red arrow illustrates the force generated by the fluidic inflation of the gripping chamber and transmitted through the screws connecting the orange and green components, which applies a concentrated load to the three-segment beam of the orange component. 
The blue component represents the gripping chamber (fluidic actuator), the green component acts as a fixed support and transmission interface, and the orange component is the deformable clamping beam whose deflection governs the reaction pressure of the system.

\subsubsection{Geometric Definition}
The clamping beam is modeled as a three-segment piecewise linear Euler--Bernoulli beam with segment lengths \(L_i\), bending stiffnesses \(E I_i\), and inclination angles \(\theta_i\) with respect to the global \(x\)-axis (\(i=1,2,3\)).
The cumulative node coordinates satisfy
\[
x_i = x_{i-1} + L_i\cos\theta_i, \qquad 
y_i = y_{i-1} + L_i\sin\theta_i,
\]
and the tip of the beam, corresponding to the point of load application, is \((X_L,Y_L)=(x_3,y_3)\).
For global-coordinate formulation, each segment is expressed as
\[
y(x) = y_{i-1} + \tan\theta_i\,(x - x_{i-1}), \qquad x\in[x_{i-1},x_i].
\]
The infinitesimal arc length is related to the global coordinate by
\[
ds = \sqrt{1+\left(\frac{dy}{dx}\right)^2}dx = \frac{dx}{\cos\theta_i}.
\]

\subsubsection{Bending Moment and Deflection}
The hydraulic pressure \(p\) generated in the chamber produces a net vertical load
\[
P = pA_{\mathrm{eff}},
\]
where \(A_{\mathrm{eff}}\) is the effective projected area transmitting the chamber pressure to the beam.
For a vertical concentrated load \(\mathbf{F} = (0,P)\) acting at the free end, the bending moment at a cross-section \(x\) on segment \(i\) is
\[
M_i(x) = P(X_L - x), \qquad x\in[x_{i-1},x_i].
\]
Using the unit-load (virtual work) method, the tip deflection in the load direction is
\begin{equation}
\delta = 
\frac{P}{E}
\sum_{i=1}^{3}
\frac{1}{I_i}
\int_{x_{i-1}}^{x_i}
\frac{(X_L - x)^2}{\cos\theta_i}\,dx.
\label{eq:deflection_global}
\end{equation}
The conversion factor \(1/\cos\theta_i\) originates from the transformation of arc-length integration \(ds\) to global coordinates.

\subsubsection{Closed-Form Solution}
Evaluating Eq.~\eqref{eq:deflection_global} gives the closed-form expression:
\begin{equation}
\boxed{
\delta
= \frac{P}{E}
\sum_{i=1}^{3}
\frac{
L_i\,\Delta x_{i0}^{\,2}
- \Delta x_{i0}\cos\theta_i\,L_i^{2}
+ \tfrac{\cos^{2}\theta_i}{3}\,L_i^{3}
}{I_i}
},
\label{eq:deflection_closed}
\end{equation}
where
\[
\Delta x_{i0} = X_L - x_{i-1}.
\]
When all segments are collinear (\(\theta_i=0\)) and \(I_i\equiv I\), Eq.~\eqref{eq:deflection_closed} reduces to the classical single-beam result \(\delta=PL^3/(3EI)\).

\subsubsection*{Effective Compliance and Stiffness}
The overall bending compliance and equivalent stiffness of the clamping beam are defined as
\[
C_b = \frac{\delta}{P}, \qquad k_b = \frac{1}{C_b}.
\]
These quantities determine how the structural deformation of the clamping fixture couples with the hydraulic response of the gripping chamber.

\subsubsection{Coupling with Chamber Pressure and Object Stiffness}
The relationship between hydraulic pressure, beam deformation, and gripping response can be summarized as follows:
\[
P = pA_{\mathrm{eff}}, \qquad 
\delta = C_b P, \qquad 
p = \frac{\delta}{A_{\mathrm{eff}} C_b}.
\]
For volume-controlled inflation, the internal pressure \(p\) depends on the chamber volume change \(\Delta V\), which determines the applied load \(P(\Delta V)\) and corresponding beam deflection \(\delta(\Delta V)\).
If an external object of stiffness \(k_o\) is grasped, the object and the beam deform in series:
\[
u = \delta_b + \delta_o = \frac{P}{k_b} + \frac{P}{k_o}, 
\qquad
k_{\mathrm{eq}} = \frac{k_b k_o}{k_b + k_o}.
\]
A stiffer object (larger \(k_o\)) restricts the beam deflection, leading to higher internal pressure and a greater reaction force.
As the inflation continues, the beam deflection approaches the snap-through distance of the chamber,
beyond which the reaction pressure reaches a stable plateau. This framework establishes a quantitative link between the chamber’s pressure evolution and the stiffness-dependent deformation of the clamping fixture.

\subsection{Fixed-Distance Grasping Experiment}

After confirming the stiffness-adaptive behavior of our gripper design, we further verified the feasibility of the proposed source-free bistable gripper by assembling it according to the procedure described in Section~\ref{subsection:mf} and performing proof-of-concept experiments using representative laboratory objects with different sizes and stiffness. This approach allows controlled validation of the gripper’s size-selective and stiffness-adaptive behavior, reflecting its potential use in centimeter-scale sampling tasks with targeted size.

The grasping experiments were divided into two groups: a small-size group (2 cm \textless gripping size \textless 4 cm) and a large-size group (8 cm \textless gripping size \textless 10 cm), demonstrating that the proposed gripper can effectively handle objects of both smaller and larger fixed sizes. The upper limit of the gripping size in each group was defined by the preset normal distance between the clamping fixtures, while the lower limit was determined by the maximum deformation displacement achievable by the gripping chambers during snap-through (1 cm) in this setup.

As shown in Fig. \ref{fig:test-result-2}a–c, for the small-size group experiments, three representative objects were tested: an effervescent tablet tube (stiff), a centrifuge tube (stiff), and a pack of biscuits (with thinner packaging that is more susceptible to deformation). 

Meanwhile, Fig. \ref{fig:test-result-2}d–f present three representative objects from the large-size group: an empty black box (stiff), an empty water-soaked paper cup (with walls and bottom that are easily deformed), and a fish oil bottle (stiff).

After the contact chamber was triggered, all objects were successfully grasped and lifted by the gripping chambers. The corresponding hydraulic pressure variations are presented in Fig. \ref{fig:test-result-2}g and h. Because the interconnected liquid circuit could not be completely purged of air and a transient pressure drop occurred during the snap-through transition between the contact and gripping chambers, the measured pressures were lower than those obtained in the quasi-static tests. Nevertheless, the results clearly show a snap-through response pressure exceeding 6 kPa (critical pressure for snap-through), after which the gripping pressure remained stable during object holding. Furthermore, the experiments demonstrate that the gripping module passively adjusts its output pressure according to the resistance encountered during the snap-through process, thereby adapting its gripping force to objects of different stiffness.

Specifically, for stiff objects such as the black box, fish oil bottle, and tubes, the gripper generated a hydraulic response of approximately 6 kPa. For the biscuit pack—with a flexible outer package but relatively firm internal contents—the response pressure was around 4 kPa. In most cases, the snap-through and grasping response occurred within one second. The case of the water-soaked paper cup was distinct: being highly deformable and fragile, it exhibited a slower response time. However, after activation, the gripper automatically adjusted to exert an extremely low internal pressure corresponding to the cup’s reduced structural stiffness (low bending stiffness of its curved surface), thereby preventing damage to the delicate object.

% Section4--------------------------------------------------------
\section{Conclusions}

This work presents a source-free fluid-driven gripper that achieves stable grasping through internal liquid redistribution among bistable snap-through chambers. By integrating a bistable mechanism into a closed hydraulic circuit, the proposed design eliminates the need for external pumps, enabling compact and energy-efficient operation. Finite element simulations and experiments demonstrate that the snap-through chambers exhibit tunable bistability governed primarily by their geometric $45^{\circ}$ tilt angle, providing optimal switching performance.

The gripper exhibits both size selectivity and passive stiffness adaptation. Under fixed-distance configurations, it successfully grasped objects of appropriate size and varying stiffness, automatically adjusting its internal hydraulic pressure (1–6 kPa range) according to object rigidity. These results confirm that the bistable fluidic architecture enables reliable, self-contained manipulation without continuous energy input.

Taken together, these promising results point to significant potential for the snap-through–inspired approach in soft robotics—enabling size-specific sampling, low-power or autonomous gripping, and adaptive manipulation—while its compact, source-free hydraulic architecture supports on-board use on mobile or untethered platforms for underwater and field deployments without external pumps.

% ACKNOWLEDGMENT-------------------------------------------------
\section*{Acknowledgements}
We thank Ji Qi for his assistance with the experiments.

\bibliographystyle{ieeetr}
\bibliography{citationlist}

\end{document}